\documentclass{article}
\usepackage{spconf,graphicx,hyperref}
\usepackage{amsmath}   
\usepackage{amssymb}   
\usepackage{amsfonts}  

\makeatletter
\def\thebibliography#1{%
  \section*{References}\vspace{-1.0\baselineskip} 
  \list{\@biblabel{\arabic{enumiv}}}{%
    \settowidth\labelwidth{\@biblabel{#1}}%
    \leftmargin\labelwidth
    \advance\leftmargin\labelsep
    \usecounter{enumiv}%
    \setlength{\topsep}{0pt}%
    \setlength{\partopsep}{0pt}%
    \setlength{\itemsep}{0pt}%
    \setlength{\parsep}{0pt}%
  }%
  \sloppy\clubpenalty4000\widowpenalty4000\sfcode`\.=1000\relax}

\makeatother

\usepackage{algorithm}
\usepackage{algorithmic}
\usepackage{titlesec}
\usepackage{arydshln}
\usepackage{multirow}
\usepackage{tabularx}
\usepackage{colortbl}
\usepackage{multirow}
\usepackage{array}
\usepackage{makecell} 
\usepackage[hang]{footmisc}
\usepackage[dvipsnames]{xcolor}
\definecolor{c1}{HTML}{FF8C00}
\definecolor{c2}{HTML}{E57373}
\definecolor{c3}{HTML}{FFB2B2}
\definecolor{c4}{HTML}{FFE3E3}
\definecolor{c5}{HTML}{FFB400}
\usepackage{textcomp} 
\usepackage{booktabs} 
\makeatletter

\renewenvironment{thebibliography}[1]{%
  \begin{oldthebibliography}{#1}%
    \setlength{\itemsep}{0pt}
    \setlength{\parskip}{0pt}
    \setlength{\parsep}{0pt}%
    \setlength{\topsep}{0pt}
    \setlength{\partopsep}{0pt}%
  }%
  {\end{oldthebibliography}}
\makeatother

\AtBeginDocument{%
  }

\title{Teacher-Guided Pseudo Supervision and Cross-Modal Alignment for Audio-Visual Video Parsing}
\name{Yaru Chen$^{1}$,
      \text{Ruohao Guo}$^{2}$, 
      \text{Liting Gao}$^{1}$, 
      Yang Xiang$^{1}$,
      Qingyu Luo$^{1}$,
      Zhenbo Li$^{3}$, 
      Wenwu Wang$^{1}$
      }
\address{$^{1}$Centre for Vision Speech and Signal Processing~(CVSSP), University of Surrey, United Kindom\\
	$^{2}$School of Intelligence Science and Technology, Peking University, China\\
    $^{3}$College of Information and Electrical Engineering, China Agricultural University, China
}
\begin{document}
%
\maketitle
\begin{abstract}
Weakly-supervised audio-visual video parsing (AVVP) seeks to detect audible, visible, and audio-visual events without temporal annotations. Previous work has emphasized refining global predictions through contrastive or collaborative learning, but neglected stable segment-level supervision and class-aware cross-modal alignment. To address this, we propose two strategies: (1) an exponential moving average (EMA)-guided pseudo supervision framework that generates reliable segment-level masks via adaptive thresholds or top-$k$ selection, offering stable temporal guidance beyond video-level labels; and 
(2) a class-aware cross-modal agreement (CMA) loss that aligns audio and visual embeddings at reliable segment–class pairs, ensuring consistency across modalities while preserving temporal structure.
Evaluations on LLP and UnAV-100 datasets shows that our method achieves state-of-the-art~(SOTA) performance across multiple metrics.

\end{abstract}
\vspace{-3mm}
\section{Introduction}
\label{avvp}
\vspace{-1mm}
Weakly-supervised audio-visual video parsing (AVVP)~\cite{tian2020unified} aims to localize audible, visible, and audio-visual events in unconstrained videos. As illustrated in Fig.~\ref{avvp}, only video-level annotations are available during training, making it highly challenging to infer precise temporal boundaries and modality-specific events. This task is critical for applications such as audio-visual understanding, event detection, and segmentation~\cite{xue2021audio,li2024boosting,guo2025audio}. Recent AVVP studies have explored strategies such as multi-instance learning for weak labels~\cite{tian2020unified,wu2021exploring}, attention mechanisms to highlight informative segments~\cite{chen2024cm,jiang2022dhhn}, and contrastive or collaborative learning to exploit audio–visual correlations~\cite{sardari2024coleaf,gao2025learning}. Although these approaches improve performance, two key challenges remain.

\begin{figure}[t] 
\includegraphics[width=0.48\textwidth]{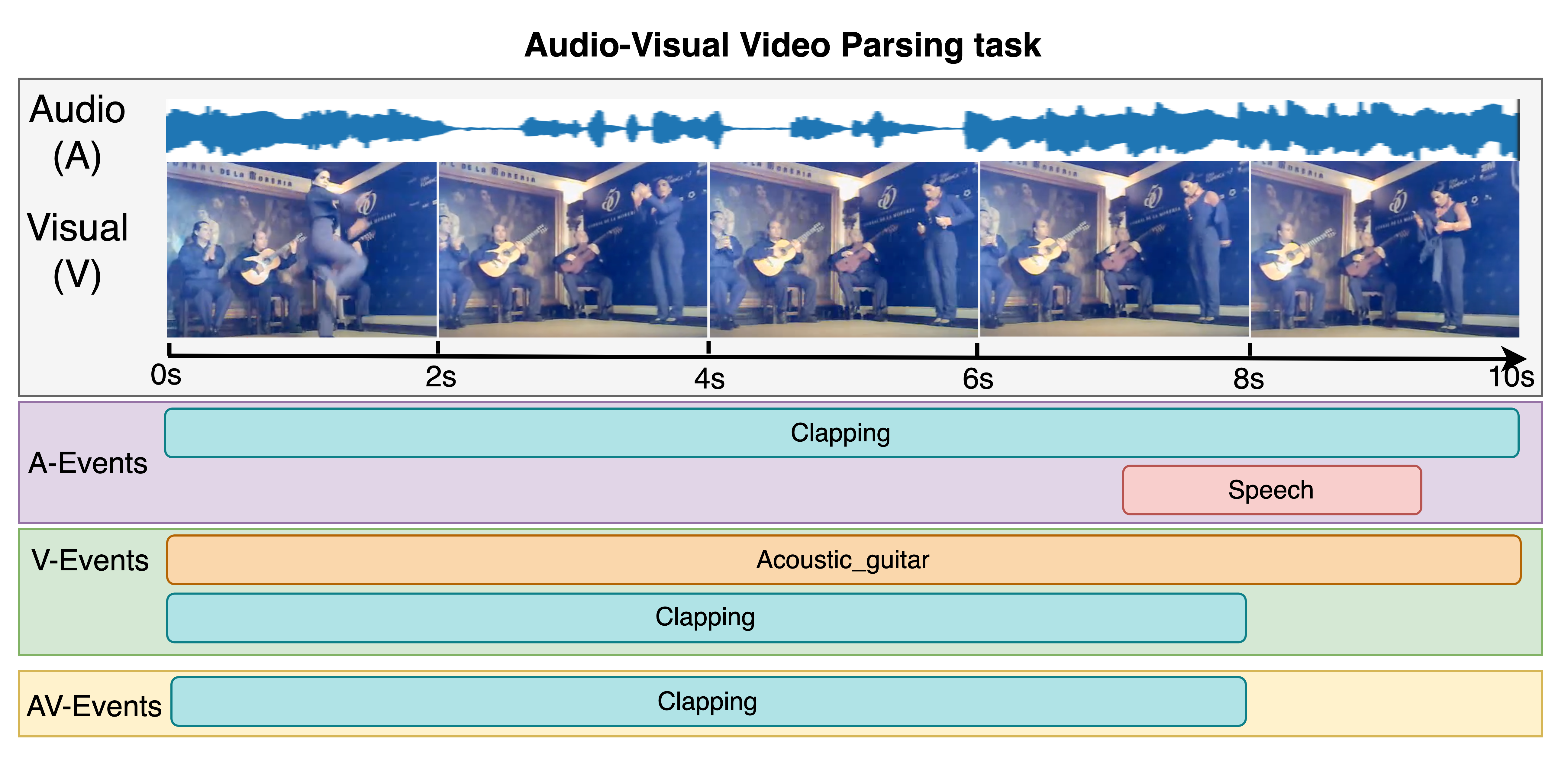} 
    \vspace{-4mm}
    \caption{Illustration of the AVVP task.} 
    \label{avvp}
\end{figure} 
First, the lack of segment-level supervision, due to training with only video-level labels, makes it difficult to achieve stable learning. Previous works often propagate coarse labels to all segments or apply simple thresholding, which introduces noise~\cite{cheng2022joint,gao2023collecting}. Although large pre-trained models (e.g., CLIP~\cite{radford2021learning}, CLAP~\cite{wu2023large}) have been used to generate pseudo labels~\cite{lai2023modality,rachavarapu2024weakly}, these are typically static and cannot be refined during training, leaving them prone to noise and domain mismatch. Thus, more reliable, dynamically updated pseudo supervision is needed. Second, most cross-modal approaches align modalities by maximizing global audio–visual similarity~\cite{sardari2024coleaf,mo2022multi}, overlooking that different classes may occur in different modalities at different times. Without class-aware, segment-level alignment, models risk forcing mismatches between unrelated events, leading to suboptimal localization.
\begin{figure*}[h]
    \centering    \includegraphics[width=0.85\textwidth]{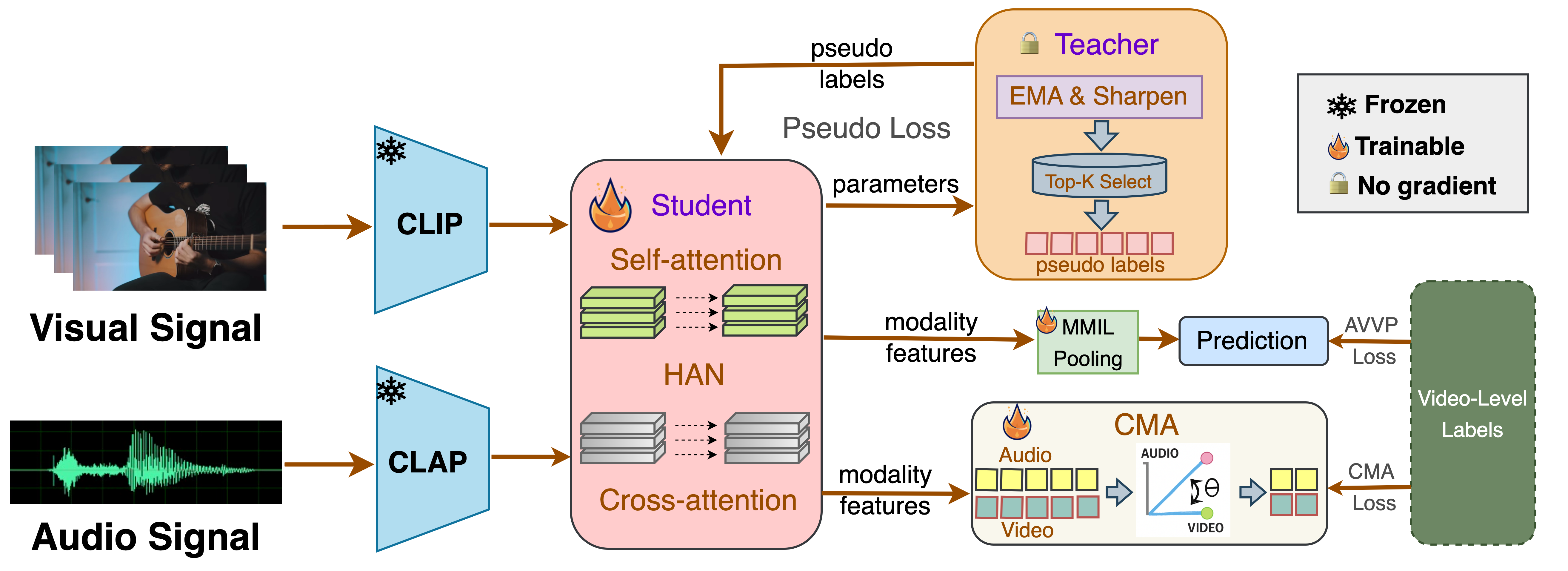} 
     \caption{Framework of E-CMANet. The EMA Teacher provides stable pseudo labels to guide the Student, and the CMA module enforces class-aware cross-modal alignment. The final loss combines Pseudo loss, AVVP loss, and CMA loss.}
    \label{ECMA}
\end{figure*}

To overcome these limitations, we proposed E-CMA, which includes two strategies. (1) Exponential moving average (EMA)-guided pseudo supervision~\cite{tarvainen2017mean}. We adopt a teacher–student framework where the teacher, updated via EMA of student parameters, periodically generates segment-level pseudo masks from frame-wise predictions using adaptive thresholds or top-k selection per class. This transforms noisy video-level labels into stable, dynamically updated supervision, reducing error propagation from static pseudo labels. (2) Class-aware cross-modal agreement (CMA) loss. Rather than enforcing global similarity between audio and visual embeddings, CMA aligns modalities only at confident temporal–class positions where both modalities strongly indicate the same event. This selective alignment prevents over-matching asynchronous content and provides fine-grained, event-consistent supervision.

We evaluated our method on the widely used AVVP benchmarks, the LLP dataset~\cite{tian2020unified} and also the weakly-labeled UnAV-100 dataset~\cite{geng2023dense}. Experimental results show that our approach achieves state-of-the-art~(SOTA) performance on multiple metrics, which highlights the importance of stable pseudo supervision and fine-grained cross-modal agreement for advancing weakly-supervised AVVP.
\vspace{-5mm}
\section{Proposed Methods}
\label{sec:format}
\vspace{-3mm}
\subsection{Problem Statement}
\vspace{-1mm}
The goal of AVVP is to determine whether the event occurring in each time segment of a video is audio-only, video-only, or audio-visual event and to determine its category. Given a $T$ second video $\mathcal{X}$ that is partitioned into $T$ non-overlapping segments, each one second long. We denote the sequence as $\mathcal{X}=\{(x_t^a,x_t^v)\}_{t=1}^T$, where $x_t^a,x_t^v\in \mathbb{R}^{d}$ represent the audio and visual features at time $t$ and $d$ represents the dimension of the features. For each segment, we define the event labels as $y_t^a\in \{0,1\}^C,y_t^v\in \{0,1\}^C,y_t^{av}\in \{0,1\}^C$, where $C$ is the number of event classes.  The audio-visual event occurs only when both modalities detect the same event at time $t$, that means $y_t^{av}=y_t^a\odot y_t^v$, where $\odot$ denotes element-wise multiplication. AVVP is often a weakly supervised task, as only video-level labels are available during training, i.e. $y\in \{0,1\}^C$. In contrast, evaluation is performed with segment-level annotations, where fine-grained event boundaries are provided for both modalities.
\vspace{-4mm}
\subsection{Framework}
\label{sec:framework}
\vspace{-1mm}
As shown in Fig.~\ref{ECMA}, our framework builds upon the CoLeaF~\cite{sardari2024coleaf} baseline. We first extract audio and visual features using pre-trained CLAP~\cite{wu2023large} and CLIP~\cite{radford2021learning} encoders. These features are refined by a Hierarchical Attention Network (HAN)~\cite{tian2020unified} with self- and cross-attention to capture intra- and inter-modal dependencies. The updated segment representations are aggregated by (Multimodal Multiple Instance Learning)~MMIL pooling~\cite{tian2020unified} for video-level predictions.
To enhance temporal supervision, we further introduce an EMA-guided teacher–student scheme. The teacher, updated by the exponential moving average, periodically generates stable segment-level pseudo masks that supplement weak video-level labels. In addition, a class-aware CMA loss is applied to enforce alignment between audio and visual embeddings in confident event segments. These novel designs enable more reliable segment-level learning and fine-grained cross-modal alignment.
\vspace{-5mm}
\subsection{Teacher-Guided Pseudo Supervision}
\vspace{-1mm}
To address the lack of segment-level annotations and the limitations of fixed, non-adaptive pseudo labels, we introduce an EMA-guided teacher–student framework, where a slowly updated teacher generates reliable segment-level masks, which are then used to supervise the training of the student network.

As shown in Fig.~\ref{ECMA}, the student network follows the architecture we introduced in Section~\ref{sec:framework}~, which includes CLAP and CLIP encoders, HAN aggregation, and MMIL pooling. It is updated by backpropagation with standard weakly-supervised objectives. The student network produces audio and visual probabilities $\hat{P}^a_t,\hat{P}^v_t \in [0,1]^C$. Then we fuse them into a joint prediction vector,
\vspace{-2mm}
\begin{equation}
\label{EMA0}
\hat{P}_t = \frac{1}{2}(\hat{P}^a_t+\hat{P}^v_t), \hat{P}_t\in[0,1]^C
\end{equation}
\vspace{-5mm}

The teacher network shares the same backbone architecture as the student network but is not optimized by gradient descent. Instead, its parameters are updated as an EMA~\cite{tarvainen2017mean} of the student’s parameters, which maintains a weighted average of the previous student parameters to form a more stable teacher model. Let $\theta$ denote the parameters of the student at iteration $k$, and $\theta'$ denote the teacher parameters, thus:
\vspace{-2mm}
\begin{equation}
\label{EMA1}
\theta'_k = \alpha \theta^{'}_{k-1}+(1-\alpha)\theta_k
\end{equation}
where $\alpha\in[0,1)$ is a momentum coefficient. This update ensures the teacher evolves smoothly, making its predictions more stable than the student network.

To provide effective temporal supervision for the student, we transform the teacher’s segment-level predictions into binary pseudo masks that indicate reliable event occurrences. The teacher’s audio and visual predictions are first fused into a joint score:
\vspace{-2mm}
\begin{equation}
\label{EMA2}
\tilde{P}_t = \frac{1}{2}(\tilde{P}^a_t+\tilde{P}^v_t), \tilde{P}_t\in[0,1]^C
\end{equation}
\vspace{-5mm}

From $\tilde{P}_t$, we can derive binary pseudo masks $M_t\in\{0,1\}^C$ in two ways: adaptive thresholding or top-$k$ selection. For adaptive thresholding, the threshold is dynamically adjusted by the mean prediction confidence for each class $c$, that means:
\begin{equation}
\label{EMA3}
\tau = \gamma \cdot \frac{1}{T}\sum^T_{t=1}\tilde{P}_{t,c}
\end{equation}
where $\gamma$ is the scaling factor, then we define the pseudo mask:
\vspace{-3mm}
\begin{equation}
\label{EMA4}
M_{t,c} =
\begin{cases}
1, & \text{if } \tilde{P}_{t,c} \geq \tau_c, \\
0, & \text{otherwise},
\end{cases}
\end{equation}
where $\mathbf{M}\in\{0,1\}^{T\times C}$ is the pseudo mask matrix of elements $M_{t,c}$, and $\tilde{P}_{t,c}$ indicate the teacher's predicted confidence score for class $c$ at segment $t$. Alternatively, wee can use $\mathrm{Top}-k$ selection to generate pseudo mask: $\text{if } t \in \mathrm{Top-}k\!\left(\{\tilde{P}_{t,c}\}_{t=1}^T,\, k\right)$,
where $\mathrm{Top-}k$ returns the inidices of the $k$ highest confidence scores among all $T$ segments for class $c$, and $k$ is a hyperparameter which is the number of selected segments. After generating $\mathbf{M}$, we integrate them into the learning objective by enforcing consistency between the student’s predictions and the pseudo labels, which we use a masked binary cross-entropy loss:
\begin{equation}
\label{EMA6}
\mathcal{L}_{\text{pseudo}} = 
\frac{1}{\lVert \mathbf{M} \rVert_{1}}
\sum_{t=1}^{T} \sum_{c=1}^{C} 
\mathbf{M}_{t,c} \, \ell \!\left( \hat{P}_{t,c}, 1 \right),
\end{equation}
where $\lVert \mathbf{M} \rVert_{1}$ is the L1 norm. By this design, only the trusted segment–class pairs indicated by $\mathbf{M}$ contribute to the loss, while the remaining uncertain positions are ignored. This prevents noise accumulation and provides consistent temporal guidance beyond video-level labels.
\vspace{-3mm}
\subsection{Class-Aware Cross-Modal Alignment Loss}
\vspace{-1mm}
Although pseudo supervision provides reliable temporal masks, it does not explicitly enforce feature-level alignment across modalities. Hence, we introduce a class-aware CMA loss, which selectively encourages audio and visual embeddings to be consistent at confident segment–class positions.

Concretely, for each time step $t\in \{1,2,...,T\}$ and event class $c\in\{1,2,...,C\}$, we select valid segment-class pairs $(t,c)$ which meet two conditions:~(1) The predicted probabilities for both modalities $\hat{P}^a_{t,c}$ and $\hat{P}^v_{t,c}$ over their respective confidence thresholds $\tau_a$ and $\tau_v$; (2) The video-level label $y_c=1$, indicating the event $c$ occurs in the video.
We denote $\Omega$ as the set of these valid pairs, and apply the CMA loss for these pairs. For each pair~$(t,c)\in\Omega$, we calculate the cosine similarity between audio and visual features: 
\vspace{-2mm}
\begin{equation}
\label{CMA0}
s_{t,c} = 
\frac{\left( x_t^a \right)^{\top} x_t^v}
{\lVert x_t^a \rVert_{2} \cdot \lVert x_t^v \rVert_{2}}
\end{equation}
Then, the CMA loss is formulated as the average cosine distance across all valid pairs:
\vspace{-2mm}
\begin{equation}
\label{CMA2}
\mathcal{L}_{\text{CMA}} = 
\frac{1}{|\Omega|}
\sum_{(t,c)\in\Omega} \left( 1 - s_{t,c} \right)
\end{equation}
\vspace{-5mm}

By restricting the loss to confident and label-consistent segment–class pairs, CMA suppresses noisy interactions and reinforces semantically meaningful alignment across modalities. Overall, the total loss $l$ for our framework is as follows:
\begin{equation}
\label{CMA2}
\mathcal{L} = \mathcal{L}_{AVVP}+\mathcal{L}_{pseudo}+\mathcal{L}_{CMA}
\end{equation}
Here, $\mathcal{L}_{AVVP}$ is the standard binary cross-entropy loss between predictions and ground-truth labels.
\vspace{-5mm}
\section{Experimental Results}
\label{Experimental Results}
\vspace{-3mm}
\subsection{Experimental Setup}
\vspace{-2mm}
\textbf{Dataset.} We evaluate our model on the LLP dataset~\cite{tian2020unified}, which is the benchmark for AVVP. It includes 11,849 10-seconds videos covering 25 event categories.
We also perform tests on UnAV-100~\cite{geng2023dense}, a large-scale dataset for audio-visual event localization, containing 10,790 videos and over 30k event instances from 100 classes. Following CoLeaF~\cite{sardari2024coleaf}, we use only video-level labels for training UnAV-100.\\
\textbf{Implementation Details.} For the LLP dataset, we extract 768-dimensional audio and visual features using the pre-trained CLAP~\cite{wu2023large} and CLIP~\cite{radford2021learning}.
For the UnAV-100 dataset, we extract 2048-dimensional visual features using a two-stream I3D model (RGB + RAFT)~\cite{teed2020raft}, and 128-dimensional audio features using pre-trained VGGish~\cite{hershey2017cnn}.\\
\textbf{Evaluation Metrics.} We evaluate our model using F1-scores for three event types: audio (A), visual (V), and audio-visual (AV). A prediction is considered correct if the intersection-over-union (IoU) with ground truth exceeds 0.5.
We computed the scores at the segment and event levels. The segment-level evaluation compares predictions and labels frame-by-frame, while event-level evaluation merges consecutive positive segments into a single event.
We further report Type@AV, the average over A, V, and AV events, and Event@AV, which assesses the overall audio-visual event detection performance in each video.
\begin{table*}[t]
  \caption{he performance of E-CMA and comparative methods in AVVP, with the best results highlighted in \textbf{bold} and the second results highlighted in \underline{text}.}
  \small
  \centering
  \label{tab:llp-results}
  \vspace{1mm}
  \renewcommand{\arraystretch}{0.8}
  \begin{tabular}{p{2cm}|c|ccccc|ccccc}
    \toprule
    \multirow{2}{*}{Model} & \multirow{2}{*}{Venue} & \multicolumn{5}{c|}{Segment-level (\%)} & \multicolumn{5}{c}{Event-level (\%)}\\
    \cmidrule{3-7} \cmidrule{8-12}
    &  & A & V & AV & Type@AV & Event@AV & A & V & AV & Type@AV & Event@AV \\
    \midrule
    HAN~\cite{tian2020unified} & ECCV'20 & 60.1 & 52.9 & 48.9 & 54.0 & 55.4 & 51.3 & 48.9 & 43.0 & 47.7 & 48.0 \\
    MGN~\cite{mo2022multi} & NeurIPS'22 & 60.8 & 55.4 & 50.0 & 55.1 & 57.6 & 52.7 & 51.8 & 44.4 & 49.9 & 50.0 \\
    MA~\cite{wu2021exploring} & CVPR'21 & 60.3 & 60.0 & 55.1 & 58.9 & 57.9 & 53.6 & 56.4 & 49.0 & 53.0 & 50.6 \\
    CMPAE~\cite{gao2023collecting} & CVPR'23 & 64.2 & 66.2 & 59.2 & 63.3 & 62.8 & 56.6 & 63.7 & 51.8 & 57.4 & 55.7 \\
    VALOR~\cite{lai2023modality} & NeurIPS'23 & 61.8 & 65.9 & 58.4 & 62.0 & 61.5 & 55.4 & 62.6 & 52.2 & 56.7 & 54.2 \\
    CoLeaF~\cite{sardari2024coleaf} & ECCV'24 & 64.2 & \underline{67.1} & 59.8 & 63.8 & 61.9 & \underline{57.1} & \underline{64.8} & 52.8 & \underline{58.2} & 55.5 \\
    PPL~\cite{rachavarapu2024weakly} & CVPR'24 & \underline{65.9} & 66.7 & \textbf{61.9} & \underline{64.8} & \underline{63.7} & \textbf{57.3} & 64.3 & \textbf{54.3} & \textbf{59.9} & \textbf{57.9} \\
    RLLD~\cite{gao2025reinforced} & CVM'25 & 62.2 & 66.7 & 59.3 & 62.7 & 62.4 & 55.7 & 63.1 & \underline{53.7} & 57.5 & 54.9 \\
    PPAE~\cite{gao2025learning} & TPAMI'25 & 64.3 & 66.6 & 59.6 & 63.5 & 63.0 & 57.0 & 64.1 & 52.5 & 57.9 & \underline{56.1} \\
    \midrule
    \multirow{2}{*}{\textbf{E-CMANet}} & \multirow{2}{*}{-} 
      & \textbf{66.1} & \textbf{69.9} & \underline{61.7} & \textbf{65.9} & \textbf{65.4} 
      & 54.5 & \textbf{66.6} & 53.5 & \underline{58.2} & 54.3 \\
    &  & \textcolor{blue}{(+0.2)} & \textcolor{blue}{(+2.8)} &  & \textcolor{blue}{(+1.1)} & \textcolor{blue}{(+1.7)} 
      &  & \textcolor{blue}{(+1.8)} &  &  &  \\
    \bottomrule
  \end{tabular}
\end{table*}
\vspace{-4mm}
\subsection{Overall Performance Analysis}
\vspace{-1mm}
Table~\ref{tab:llp-results} and~\ref{UnAV100} shows the comparative experimental results on the LLP and UnAV-100 datasets between our approach and previous SOTA methods. The results indicate that our methods delivers superior performance on several key metrics, with particularly notable improvements in segment-level parsing, substantially outperforming existing approaches. Since our method does not incorporate text embeddings with audio and visual features, we restrict the comparison to approaches which also do not employ such fusion.
\begin{table}[ht]
    \caption{Comparison of E-CMANet performance on the weakly-labeled UnAV-100 dataset.}
    \label{UnAV100}
    \small
    \vspace{1mm}
\centering
\begin{tabular}{lcc}
\toprule
\textbf{Method} & \textbf{AV (Seg)} & \textbf{AV (Evn)} \\
\midrule
HAN~\cite{tian2020unified}     & 35.0 & 41.4 \\
MA~\cite{wu2021exploring}       & 37.9 & 44.8 \\
JoMoLD~\cite{cheng2022joint} & 36.4 & 41.2 \\
CMPAE~\cite{gao2023collecting} & 39.7 & 43.8 \\
CoLeaF\cite{sardari2024coleaf} & 41.5 & \textbf{47.8} \\
\midrule
\textbf{E-CMANet} & \textbf{41.8} \textcolor{blue}{\textbf{(+0.3)}} & \underline{47.4}\\
\bottomrule
\end{tabular}
\end{table}

As shown in Table~\ref{tab:llp-results}, E-CMANet achieves the best overall performance on the LLP dataset. At the segment level, it consistently outperforms previous approaches, reaching 66.1\% and 69.9\% on the audio and visual modalities, both new SOTA results. It also yields improvements on joint AV metrics, with 61.7\% on AV and 65.4\% on Event@AV. At the event level, E-CMANet attains the highest visual score of 66.6\% and competitive results across other metrics, including 58.2\% on Type@AV and 54.3\% on Event@AV. These results demonstrate that E-CMANet not only enhances unimodal performance but also achieves more consistent cross-modal event parsing. On the UnAV-100 dataset with weakly supervised labels, E-CMANet achieves 41.8\% on AV (Seg), surpassing CoLeaF by +0.3\%. For AV (Event), our model obtains 47.4\%, which is competitive with the best baseline (47.8\%). These results show that E-CMANet not only enhances segment-level localization but remains strong event-level parsing ability under weak supervision.
\vspace{-4mm}
\subsection{Ablation Study}
\vspace{-2mm}
To assess the impact of the EMA and CMA modules, we conducted ablation study on the LLP dataset by removing them from our method. For fairness, we train the CoLeaF with the same feature extractor as our method, denoted as CoLeaF$\dagger$.

As shown in Table~\ref{tab:abs-results}, both modules contribute to the effectiveness of our framework. Specifically, removing CMA leads to drops in visual and audio-visual metrics at both the segment and event levels, indicating that CMA is crucial for enhancing cross-modal alignment. On the other hand, excluding EMA mainly affects event-level results, with a decrease in Event@AV, which confirms its role in capturing event-level consistency. When both modules are integrated, E-CMANet achieves the best overall performance across all metrics, surpassing the strong baseline CoLeaF and demonstrating the complementary benefits of CMA and EMA.
\vspace{-6mm}
\begin{table}[h]
  \caption{Ablation study for E-CMANet. w/o~CMA and w/o~EMA mean without CMA and EMA, respectively.}
  \vspace{1mm}
  \small
  \centering
  \label{tab:abs-results}
  \setlength{\tabcolsep}{1mm}{
  \renewcommand{\arraystretch}{0.6}
  \begin{tabular}{p{1cm}clccccc}
    \toprule
    & Method & A & V & AV & Type@AV & Event@AV\\
    \midrule
   \multirow{4}{*}{\makecell{Segment\\level}} & CoLeaF$^\dagger$ & 64.2  & 67.4  & 59.9 & 63.8 & 63.3\\
   & \textit{ w/o}~CMA & 65.4 & 68.2 & 60.4 & 64.7 & 64.4 \\
   & \textit{ w/o}~EMA & 65.9 & 68.8 & 61.0 & 65.2 & 64.8 \\
    \cmidrule{2-7}
    & \textbf{E-CMANet} & 66.1 & 69.9 & 61.7 & 65.9 & 65.4\\
    \midrule
   & Method & A & V & AV & Type@AV & Event@AV\\
    \midrule
   \multirow{4}{*}{\makecell{Event\\level}}  & CoLeaF$^\dagger$ & 53.2 & 64.1 & 52.4 & 56.6 & 52.7 \\
    & \textit{ w/o}~CMA & 54.4 & 64.7 & 52.3 & 57.2 & 53.8\\
    & \textit{ w/o}~EMA & 54.5 & 65.5 & 52.9 & 57.7 & 54.0 \\
   \cmidrule{2-7}
   & \textbf{E-CMANet} & 54.5 & 66.6 & 53.5 & 58.2 & 54.3\\
  \bottomrule
\end{tabular}
}
\end{table}
\vspace{-9mm}
\section{Conclusion}
\vspace{-3mm}
In this paper, we have presented E-CMANet, a novel framework for audio-visual video parsing that incorporates the cross-modal alignment and exponential moving average modules. EMA module establishes a teacher–student scheme, where the EMA teacher generates reliable segment-level pseudo labels to guide the student. CMA enforces class-aware cross-modal consistency at the segment level, enhancing audio–visual alignment. Experiments on LLP and UnAV-100 show the effectiveness of our framework. Our approach still relies on fixed strategies for pseudo label generation, which may not fully adapt to varying event distributions. In future work, we aim to develop more adaptive teacher–student update and selection mechanisms to tackle this issue.


\vfill
\vfill\pagebreak


\vspace{-0.5em}
\bibliographystyle{IEEEbib}

\bibliography{strings,refs}

\begin{thebibliography}{10}

\bibitem{tian2020unified}
Yapeng Tian, Dingzeyu Li, and Chenliang Xu,
\newblock ``Unified multisensory perception: Weakly-supervised audio-visual video parsing,''
\newblock in {\em European Conference on Computer Vision}. Springer, 2020, pp. 436--454.

\bibitem{xue2021audio}
Cheng Xue, Xionghu Zhong, Minjie Cai, Hao Chen, and Wenwu Wang,
\newblock ``Audio-visual event localization by learning spatial and semantic co-attention,''
\newblock {\em IEEE Transactions on Multimedia}, vol. 25, pp. 418--429, 2021.

\bibitem{li2024boosting}
Guangyao Li, Henghui Du, and Di~Hu,
\newblock ``Boosting audio visual question answering via key semantic-aware cues,''
\newblock in {\em Proceedings of the 32nd ACM International Conference on Multimedia}, 2024, pp. 5997--6005.

\bibitem{guo2025audio}
Ruohao Guo, Xianghua Ying, Yaru Chen, Dantong Niu, Guangyao Li, Liao Qu, Yanyu Qi, Jinxing Zhou, Bowei Xing, Wenzhen Yue, et~al.,
\newblock ``Audio-visual instance segmentation,''
\newblock in {\em Proceedings of the Computer Vision and Pattern Recognition Conference}, 2025, pp. 13550--13560.

\bibitem{wu2021exploring}
Yu~Wu and Yi~Yang,
\newblock ``Exploring heterogeneous clues for weakly-supervised audio-visual video parsing,''
\newblock in {\em Proceedings of the IEEE/CVF Conference on Computer Vision and Pattern Recognition}, 2021, pp. 1326--1335.

\bibitem{chen2024cm}
Yaru Chen, Ruohao Guo, Xubo Liu, Peipei Wu, Guangyao Li, Zhenbo Li, and Wenwu Wang,
\newblock ``Cm-pie: Cross-modal perception for interactive-enhanced audio-visual video parsing,''
\newblock in {\em ICASSP 2024-2024 IEEE International Conference on Acoustics, Speech and Signal Processing (ICASSP)}. IEEE, 2024, pp. 8421--8425.

\bibitem{jiang2022dhhn}
Xun Jiang, Xing Xu, Zhiguo Chen, Jingran Zhang, Jingkuan Song, Fumin Shen, Huimin Lu, and Heng~Tao Shen,
\newblock ``Dhhn: Dual hierarchical hybrid network for weakly-supervised audio-visual video parsing,''
\newblock in {\em Proceedings of the 30th ACM International Conference on Multimedia}, 2022, pp. 719--727.

\bibitem{sardari2024coleaf}
Faegheh Sardari, Armin Mustafa, Philip~JB Jackson, and Adrian Hilton,
\newblock ``Coleaf: A contrastive-collaborative learning framework for weakly supervised audio-visual video parsing,''
\newblock in {\em European Conference on Computer Vision}. Springer, 2024, pp. 1--17.

\bibitem{gao2025learning}
Junyu Gao, Mengyuan Chen, and Changsheng Xu,
\newblock ``Learning probabilistic presence-absence evidence for weakly-supervised audio-visual event perception,''
\newblock {\em IEEE Transactions on Pattern Analysis and Machine Intelligence}, 2025.

\bibitem{cheng2022joint}
Haoyue Cheng, Zhaoyang Liu, Hang Zhou, Chen Qian, Wayne Wu, and Limin Wang,
\newblock ``Joint-modal label denoising for weakly-supervised audio-visual video parsing,''
\newblock in {\em European Conference on Computer Vision}. Springer, 2022, pp. 431--448.

\bibitem{gao2023collecting}
Junyu Gao, Mengyuan Chen, and Changsheng Xu,
\newblock ``Collecting cross-modal presence-absence evidence for weakly-supervised audio-visual event perception,''
\newblock in {\em Proceedings of the IEEE/CVF conference on computer vision and pattern recognition}, 2023, pp. 18827--18836.

\bibitem{radford2021learning}
Alec Radford, Jong~Wook Kim, Chris Hallacy, Aditya Ramesh, Gabriel Goh, Sandhini Agarwal, Girish Sastry, Amanda Askell, Pamela Mishkin, Jack Clark, et~al.,
\newblock ``Learning transferable visual models from natural language supervision,''
\newblock in {\em International conference on machine learning}. PMLR, 2021, pp. 8748--8763.

\bibitem{wu2023large}
Yusong Wu, Ke~Chen, Tianyu Zhang, Yuchen Hui, Taylor Berg-Kirkpatrick, and Shlomo Dubnov,
\newblock ``Large-scale contrastive language-audio pretraining with feature fusion and keyword-to-caption augmentation,''
\newblock in {\em ICASSP 2023-2023 IEEE International Conference on Acoustics, Speech and Signal Processing (ICASSP)}. IEEE, 2023, pp. 1--5.

\bibitem{lai2023modality}
Yung-Hsuan Lai, Yen-Chun Chen, and Frank Wang,
\newblock ``Modality-independent teachers meet weakly-supervised audio-visual event parser,''
\newblock {\em Advances in Neural Information Processing systems}, vol. 36, pp. 73633--73651, 2023.

\bibitem{rachavarapu2024weakly}
Kranthi~Kumar Rachavarapu, Kalyan Ramakrishnan, et~al.,
\newblock ``Weakly-supervised audio-visual video parsing with prototype-based pseudo-labeling,''
\newblock in {\em Proceedings of the IEEE/CVF Conference on Computer Vision and Pattern Recognition}, 2024, pp. 18952--18962.

\bibitem{mo2022multi}
Shentong Mo and Yapeng Tian,
\newblock ``Multi-modal grouping network for weakly-supervised audio-visual video parsing,''
\newblock {\em Advances in Neural Information Processing Systems}, vol. 35, pp. 34722--34733, 2022.

\bibitem{tarvainen2017mean}
Antti Tarvainen and Harri Valpola,
\newblock ``Mean teachers are better role models: Weight-averaged consistency targets improve semi-supervised deep learning results,''
\newblock {\em Advances in neural information processing systems}, vol. 30, 2017.

\bibitem{geng2023dense}
Tiantian Geng, Teng Wang, Jinming Duan, Runmin Cong, and Feng Zheng,
\newblock ``Dense-localizing audio-visual events in untrimmed videos: A large-scale benchmark and baseline,''
\newblock in {\em Proceedings of the IEEE/CVF Conference on Computer Vision and Pattern Recognition}, 2023, pp. 22942--22951.

\bibitem{teed2020raft}
Zachary Teed and Jia Deng,
\newblock ``Raft: Recurrent all-pairs field transforms for optical flow,''
\newblock in {\em European conference on computer vision}. Springer, 2020, pp. 402--419.

\bibitem{hershey2017cnn}
Shawn Hershey, Sourish Chaudhuri, Daniel~PW Ellis, Jort~F Gemmeke, Aren Jansen, R~Channing Moore, Manoj Plakal, Devin Platt, Rif~A Saurous, Bryan Seybold, et~al.,
\newblock ``Cnn architectures for large-scale audio classification,''
\newblock in {\em 2017 ieee international conference on acoustics, speech and signal processing (icassp)}. IEEE, 2017, pp. 131--135.

\bibitem{gao2025reinforced}
Yongbiao Gao, Xiangcheng Sun, Guohua Lv, Deng Yu, and Sijiu Niu,
\newblock ``Reinforced label denoising for weakly-supervised audio-visual video parsing,''
\newblock in {\em International Conference on Computational Visual Media}. Springer, 2025, pp. 107--124.

\end{thebibliography}

\end{document}